\documentclass[11pt]{article}

\usepackage[preprint]{acl}
\usepackage{times}
\usepackage{latexsym}
\usepackage[T1]{fontenc}
\usepackage[utf8]{inputenc}
\usepackage{microtype}
\usepackage{graphicx}
\usepackage{booktabs}
\usepackage{multirow}
\usepackage{amsmath}
\usepackage{amssymb}
\usepackage{array}
\usepackage{xcolor}
\usepackage[nameinlink]{cleveref}
\Crefname{subsection}{Section}{Sections}

\newcommand{\model}{\textsc{ConvexLlm}}
\newcommand{\cgate}{\textsc{C-Gate}}

\newcommand{\Emb}{\mathbf{E}}
\newcommand{\tablebodyfont}{\fontsize{9pt}{10pt}\selectfont}

\title{\model{}: Scaling Convex-Hull Speech Interfaces to General-Purpose Audio Intelligence}

\author{%
  Ming-Hao Hsu$^{1,\dagger}$ \quad
  Shujie Liu$^{3,*}$ \quad
  Yuxuan Hu$^{2}$ \quad
  Yuxiang Wang$^{1}$ \\
  \bfseries Yingda Shen$^{1}$ \quad
  Jinyu Li$^{2}$ \quad
  Zhizheng Wu$^{1,*}$
}

\begin{document}
\maketitle
\begingroup
\renewcommand{\thefootnote}{}
\footnotetext{$^{1}$School of Data Science, The Chinese University of Hong Kong, Shenzhen, China. $^{2}$Microsoft Research, Redmond, WA, USA. $^{3}$Microsoft Research Asia, Hong Kong, China. $^*$Correspondence to: Shujie Liu $<$\texttt{shujliu@microsoft.com}$>$ and Zhizheng Wu $<$\texttt{wuzhizheng@cuhk.edu.cn}$>$. $^\dagger$Work done during an internship at Microsoft.}
\endgroup

\begin{abstract}
Speech and audio language models connect continuous acoustic representations to decoders pretrained primarily on text data.
Most such models use MLP or Q-Former interfaces to map acoustic features into the decoder's input space, requiring the decoder to learn how to interpret the resulting representations.
We investigate whether grounding these representations in the decoder's existing text embeddings can help it use acoustic information while better preserving its pretrained text capabilities.
We present \model{}, a general-purpose speech and audio language model whose \cgate{} interface represents audio as continuous convex combinations of frozen text embeddings.
\model{} achieves an eleven-task VoiceBench aggregate of 75.36, the highest among the open-weight models we compare, and an MMAU Test-mini accuracy of 62.50\%.
Its largest margins over the baselines are on spoken knowledge and reasoning tasks, while the strongest baselines are more accurate on audio understanding and transcription.
In a comparison with a 2B-parameter decoder, \cgate{} achieves higher aggregate spoken-language performance than MLP and Q-Former interfaces, with text-task performance close to MLP and task-dependent tradeoffs in transcription and audio understanding.
These results support convex combinations of text embeddings as a viable interface for general-purpose speech and audio language models.
\end{abstract}

\section{Introduction}
\label{sec:introduction}

Recent open-weight language models have demonstrated strong capabilities in reasoning, code generation, and instruction following \citep{deepseekai2025r1,yang2025qwen3}.
Multimodal systems extend these capabilities to audio and other modalities \citep{xu2025qwen25omni,xu2025qwen3omni}.
For audio, this extension requires the model to understand what is being said and interpret paralinguistic cues, environmental sounds, or music \citep{wang2026evoaudio}.
Spoken requests draw on the language model's knowledge and instruction-following capabilities and on information that is absent from the transcript, such as a speaker's tone or background sounds.
Speech and audio language models bring these capabilities together by connecting acoustic encoders to pretrained language decoders \citep{tang2024salmonn,chu2024qwen2audio,goel2025audioflamingo3}.
Their interface must carry diverse acoustic information in a form the decoder can learn to interpret.

\begin{figure}[t]
\centering
\includegraphics[width=\columnwidth]{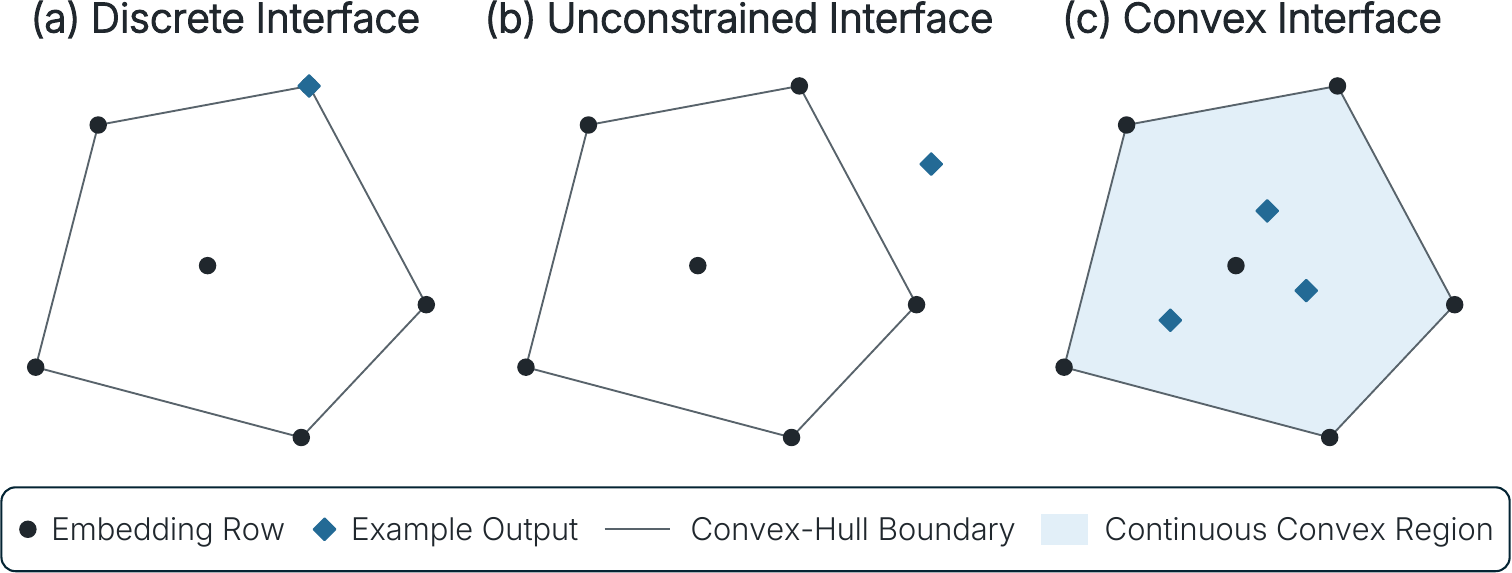}
\caption{\textbf{Comparison of Input Interfaces.}
(a) Discrete outputs select individual embedding rows.
(b) Unconstrained outputs can lie outside the convex hull of the embedding rows.
(c) Convex mixtures allow continuous variation while remaining inside the hull.}
\label{fig:convex-llm}
\end{figure}

Existing interfaces, such as MLP projectors and Q-Formers \citep{li2023blip2}, map the acoustic encoder's representations into the decoder's input space \citep{tang2024salmonn,chu2024qwen2audio}.
Vocabulary-based methods give the mapping a more explicit relation to text embeddings, for example by using speech-recognition posteriors to mix embedding rows or by introducing a text-initialized quantization codebook \citep{ma2025legoslm,yang2025softdiscretizing}.
A vocabulary-based interface can also produce continuous representations by forming weighted combinations of text embeddings.
Keeping these embeddings fixed while learning how to combine them lets the interface learn from speech while the pretrained text embeddings remain its reference.
When the weights are non-negative and sum to one, every output is a convex combination that lies inside the convex hull of the text embeddings.
Figure~\ref{fig:convex-llm} contrasts this convex interface with discrete and unconstrained continuous interfaces.

The Convex Gate (\cgate{}) is an interface of this kind \citep{hsu2026cgate}.
At each audio position, it selects a small set of frozen text embeddings and outputs their convex combination.
The original \cgate{} study established this interface in controlled speech-recognition and emotion-recognition settings.
We test the same interface in a complete model, across the broader demands of speech and audio understanding.

The central question is whether a convex interface can support general-purpose audio capability.
A fixed embedding table gives the interface a stable reference throughout training, and the decoder is trained jointly to use it.
Recognition, spoken reasoning, instruction following, and non-speech audio understanding place different demands on this interface.
We therefore evaluate these tasks together, and we compare \cgate{} with MLP and Q-Former interfaces under shared Stage 3 data, objectives, and effective-batch optimization settings.

We introduce \model{}, a speech and audio language model built on a convex input interface.
The model combines an acoustic encoder, \cgate{}, and a pretrained language decoder, and is adapted to speech and non-speech audio through staged multitask training.
We evaluate \model{} on spoken interaction, instruction following, speech recognition, and sound, music, and speech understanding.
On VoiceBench, \model{} achieves state-of-the-art performance among open-weight audio and omni models of comparable scale, with a higher eleven-task aggregate than MiniCPM-o 4.5 \citep{cui2026minicpmo45}, Kimi-Audio \citep{kimiaudio2025}, and MiMo-Audio \citep{mimoaudio2025}.
Its adaptation uses fewer than 0.3M hours of audio, more than an order of magnitude below the training scales reported for Kimi-Audio and MiMo-Audio (Appendix~\ref{app:training-scale}).
In the interface comparison, \cgate{} gives the strongest spoken-language performance with text-task performance close to MLP, while transcription and non-speech audio tasks favor different interfaces.

Our contributions are threefold.
First, we extend \cgate{} to \model{}, a general-purpose speech and audio language model built on a convex interface to frozen text embeddings.
Second, we show that \model{} achieves state-of-the-art VoiceBench performance among open-weight audio and omni models of comparable scale, driven by spoken knowledge and reasoning.
Third, we compare \cgate{}, MLP, and Q-Former interfaces under a shared Stage 3 recipe and identify distinct tradeoffs among spoken-language performance, text-task performance, transcription, and audio understanding.

\section{Related Work}
\label{sec:related-work}

\paragraph{Audio Language Models.}
Open-weight speech and audio language models such as Kimi-Audio \citep{kimiaudio2025}, MiMo-Audio \citep{mimoaudio2025}, Qwen2.5-Omni \citep{xu2025qwen25omni}, and Audio Flamingo 3 \citep{goel2025audioflamingo3} combine audio encoders or tokenizers with pretrained language decoders to support speech and audio tasks.
Learned projectors \citep{goel2025audioflamingo3} and patch encoders \citep{mimoaudio2025} map acoustic features or audio tokens to decoder inputs.
Multitask training \citep{kimiaudio2025,xu2025qwen25omni} supports recognition, instruction following, and audio understanding within a shared model.
However, unconstrained audio interfaces can produce representations far from the text embeddings, which the decoder must then learn to interpret.
Adapting LLMs to speech can also degrade their existing text capabilities \citep{cuervo2025salad}.
This motivates approaches that enable LLMs to acquire speech capabilities while preserving their existing text capabilities through constraints on the speech input representations.

\paragraph{Language Capability Transfer and Retention.}
Speech language models often underperform on spoken inputs relative to equivalent text inputs in language understanding and reasoning tasks \citep{cuervo2025salad,wang2026tars,hsu2026geometric,hsu2026entity,hsu2026affirm}.
Distillation-based approaches seek to narrow this gap by aligning speech-conditioned outputs with those produced from corresponding text inputs \citep{held2024diva,wang2024blspkd,cuervo2025salad}.
Other approaches explore simulated speech-recognition posteriors derived from text to train alignment interfaces \citep{peng2025tasu} or text-referenced reinforcement learning rewards to align speech and text reasoning \citep{wang2026tars}.
To preserve existing text capabilities during audio adaptation, training strategies also incorporate large-scale text data and subsequent text-only reinforcement learning \citep{kong2026audex}.
These approaches act through training data and supervision.
We instead study the input interface, constraining speech representations to the convex hull of frozen text embeddings from the outset of training.
We measure both text performance after adaptation and the gap between spoken and written versions of the same tasks.

\paragraph{Vocabulary-Based Interfaces.}
Vocabulary-based interfaces construct audio representations from text embeddings or codebooks initialized with them.
LegoSLM \citep{ma2025legoslm} uses connectionist temporal classification (CTC) posteriors over the model's vocabulary to form weighted sums of input embeddings.
\citet{yang2025softdiscretizing} construct mixtures from a text-initialized codebook that is updated to incorporate acoustic information.
\cgate{} \citep{hsu2026cgate} represents speech as convex mixtures of frozen text embeddings, preserving a fixed vocabulary reference while allowing continuous representations.
However, these studies have not demonstrated a general-purpose model that supports a broad range of speech and audio tasks.
Building on \cgate{}, we develop \model{} to extend this interface to general-purpose speech and audio modeling, covering speech recognition, spoken reasoning, instruction following, and non-speech audio understanding.
We also evaluate text performance after adaptation to examine how this interface supports language capability retention.

\section{\model{}}
\label{sec:method}

\ifdefined\NoArchitectureFigure
\refstepcounter{figure}\label{fig:convexllm-architecture}
\else
\begin{figure*}[!t]
\centering
\includegraphics[width=\textwidth]{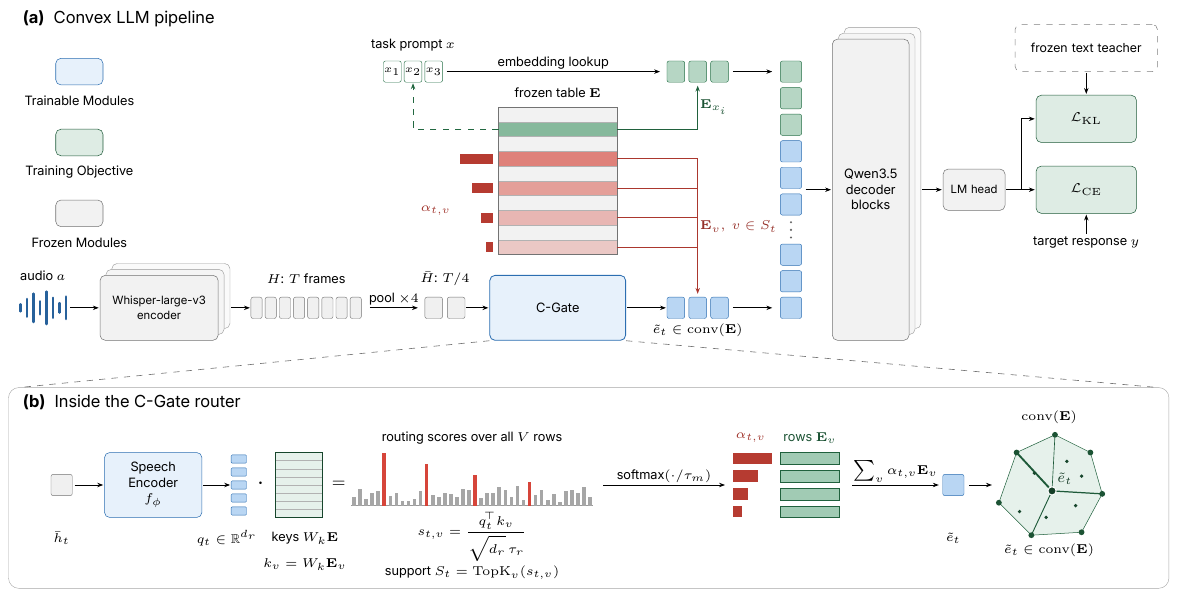}
\caption{\textbf{\model{} Architecture.}
(a) The acoustic encoder produces representations that \cgate{} maps into continuous inputs for the language decoder, which reads them next to the embedded task prompt and generates the response.
(b) For each pooled audio frame, the query projection $f_\phi$ produces a query, the router scores the vocabulary keys, selects the top-$K$ embeddings, and outputs a convex mixture of their frozen values with non-negative weights that sum to one.
The losses $\mathcal{L}_{\mathrm{CE}}$ and $\mathcal{L}_{\mathrm{KL}}$ are defined in \Cref{sec:decoder-objective}.
Gray marks the modules that are frozen in Stage 1, and the acoustic encoder and decoder blocks are trained in later stages, while the text embedding table and language-modeling head stay frozen throughout.}
\label{fig:convexllm-architecture}
\end{figure*}
\fi

\model{} is a vocabulary-based speech and audio language model, designed to adapt to audio while limiting the degradation of the decoder's text capabilities during training.
Figure~\ref{fig:convexllm-architecture} illustrates the architecture.
\Cref{sec:problem-formulation} defines the interface between the acoustic encoder and the language decoder.
\Cref{sec:convex-routing} describes how \cgate{} builds each decoder input as a convex mixture of text embeddings.
\Cref{sec:decoder-objective} defines the training objective.

\subsection{Problem Formulation}
\label{sec:problem-formulation}

\model{} encodes audio as continuous mixtures of frozen text embeddings and learns to generate a textual response from the resulting sequence.
Given a waveform $a$ and a task prompt $x$, an acoustic encoder produces states $H\in\mathbb{R}^{T\times d_a}$.
Temporal mean pooling yields $\bar H\in\mathbb{R}^{T'\times d_a}$, and the interface maps each pooled state $\bar h_t$ to a decoder input $\tilde e_t\in\mathbb{R}^{d_l}$.
Let $\Emb\in\mathbb{R}^{V\times d_l}$ be the decoder's pretrained text input-embedding table.
The table stays frozen throughout training, so learning changes how its rows are combined and interpreted while the rows themselves remain a fixed reference.

\subsection{Convex Routing}
\label{sec:convex-routing}

We use the convex-mixture construction of \cgate{} \citep{hsu2026cgate}.
\cgate{} works like a single attention lookup over the vocabulary, in which each audio position issues a query, every text embedding has a key, and the output mixes the embeddings whose keys best match the query.
We call the scoring and selection step routing and the module that performs it the router.
A query projection $f_\phi$ maps each pooled state to a query $q_t=f_\phi(\bar h_t)\in\mathbb{R}^{d_r}$, using an affine transformation followed by layer normalization and training-time dropout.
A learned linear map $W_k$ projects embedding rows to keys $k_v=W_k\Emb_v\in\mathbb{R}^{d_r}$.

For each audio position, the router scores all $V$ text embeddings by query--key similarity and keeps the $K$ highest-scoring ones:
\begin{equation}
s_{t,v} = \frac{q_t^\top k_v}{\sqrt{d_r}\,\tau_r},
\qquad S_t = \operatorname{TopK}_{v}(s_{t,v}),
\label{eq:routing}
\end{equation}
where $\tau_r>0$ is a learned routing temperature and $\operatorname{TopK}$ returns the indices of the $K$ largest scores.
A softmax over this support, sharpened by a fixed mixture temperature $\tau_m>0$, gives the mixture coefficients:
\begin{equation}
\alpha_{t,v} = \frac{\exp(s_{t,v}/\tau_m)}{\sum_{u\in S_t}\exp(s_{t,u}/\tau_m)},\quad v\in S_t.
\label{eq:mixture-weights}
\end{equation}
Selection depends only on the ordering of the scores, so both temperatures act on the sharpness of the mixture alone.
Training adjusts this sharpness through $\tau_r$, and $\tau_m$ rescales it by a fixed factor.
The interface output is the weighted sum of the selected text embeddings:
\begin{equation}
\tilde e_t = \sum_{v\in S_t}\alpha_{t,v}\Emb_v.
\label{eq:cgate}
\end{equation}
Because the mixture coefficients are non-negative and sum to one, each interface output lies within the convex hull of the frozen text embeddings.
The selection itself is not differentiated, and gradients reach the router through the scores of the selected embeddings.

\subsection{Decoder Conditioning and Training Objective}
\label{sec:decoder-objective}

The interface outputs fill the audio positions of the decoder input next to the embedded task prompt, and the decoder generates the textual response autoregressively.
We train the model with teacher-forced cross-entropy on the target response.
For an audio input $a$, a task prompt $x$, and a response $y_{1:N}$, the token-level loss for model parameters $\theta$ is
\begin{equation}
\mathcal{L}_{\mathrm{CE}}(a,x,y_i)
= -\log p_\theta(y_i \mid a,x,y_{<i}).
\label{eq:target-loss}
\end{equation}
Only response tokens contribute to the loss, so prompt tokens, audio positions, and padding are excluded.
Text-only training examples use the same objective without audio.

\noindent\textbf{Text-Teacher Distillation.}
For the main model, audio examples whose spoken content is also available as text are trained by distillation from the text decoder instead of cross-entropy.
A frozen copy of the original decoder, the text teacher, reads the written form $w$ of the spoken input, namely the written question for a synthesized instruction or the transcript for recorded speech.
The audio-conditioned model is trained to match the teacher's next-token distribution at every response position:
\begin{equation}
\begin{split}
\mathcal{L}_{\mathrm{KL}}(a,x,w,y_i)
= \mathrm{KL}\big(&p_{\bar\theta}(\cdot \mid w,y_{<i})\,\big\|\\
&p_\theta(\cdot \mid a,x,y_{<i})\big),
\end{split}
\label{eq:distill-loss}
\end{equation}
where $\bar\theta$ denotes the frozen teacher parameters.
The teacher reads the content that the audio carries, so this loss asks the model to respond to speech as the original decoder responds to the same request in text.
Audio examples without a written counterpart and text-only examples keep the cross-entropy loss.
The teacher is used only during training, and inference runs the audio-conditioned model alone.

\section{Experimental Setup}
\label{sec:experimental-setup}

We train our model on a mixture of public speech and audio datasets and synthetically generated spoken instruction and reasoning data.
We evaluate the trained model on representative speech and audio benchmarks and compare its performance with other open-weight speech and audio language models.

\subsection{Models and Implementation Details}

We use Whisper-large-v3 \citep{openai2023whisperlargev3} as the acoustic encoder.
For the language decoder, we use the Qwen3.5 series \citep{qwen2026qwen35}, with Qwen3.5-9B for the main model and Qwen3.5-2B for the controlled comparisons of interfaces and support sizes.
We apply temporal mean pooling to the encoder outputs with a stride of four.
\cgate{} uses a routing dimension of $d_r=512$ and a support size of $K=16$.
The routing temperature $\tau_r$ is initialized to 1.0, the mixture temperature $\tau_m$ is 0.7, and dropout in the query projection is 0.1.

\subsection{Multi-Stage Training}
\label{sec:staged-adaptation}

We train in three stages.
Stage 1 trains only the interface to establish an initial alignment between audio representations and the decoder, keeping the encoder and decoder frozen.
Stage 2 also trains the encoder so that its representations better suit the downstream tasks and the decoder.
Stage 3 also trains the decoder blocks, so the encoder, interface, and decoder blocks are fine-tuned jointly.
The text input-embedding table and language-modeling head remain frozen throughout all three stages.
The 9B model applies the distillation loss of Equation~\ref{eq:distill-loss} in Stage 3 (Appendix~\ref{app:main-training}).
The 2B comparisons use cross-entropy only and are evaluated after 18,000 Stage 3 updates.

\subsection{Training Data, Baseline Models, and Benchmarks}

\subsubsection{Training Data}

We train on public speech recognition datasets, including MLS English \citep{pratap2020mls}, GigaSpeech \citep{chen2021gigaspeech}, LibriSpeech \citep{panayotov2015librispeech}, Emilia \citep{he2024emilia}, Common Voice English \citep{ardila2020commonvoice}, and FLEURS English \citep{conneau2022fleurs}.
We also include audio captioning and understanding datasets, including WavCaps \citep{mei2024wavcaps}, Clotho \citep{drossos2020clotho}, and MusicCaps \citep{agostinelli2023musiclm}, together with synthetic spoken instruction and reasoning examples.

These data provide multitask supervision for speech recognition, spoken question answering, reasoning, dialogue, instruction following, and non-speech audio understanding.
We also include a small proportion of text-only examples, which we call text replay, to help preserve the decoder's text capabilities.
Appendix~\ref{app:data-audit} describes the data sources and the sampling proportions of the 2B comparison.

\subsubsection{Baseline Models}

We compare \model{} with representative open-weight speech and audio language models and omni models, including Qwen2.5-Omni \citep{xu2025qwen25omni}, Qwen2-Audio \citep{chu2024qwen2audio}, Kimi-Audio \citep{kimiaudio2025}, Audio Flamingo 3 \citep{goel2025audioflamingo3}, MiMo-Audio \citep{mimoaudio2025}, WavLLM \citep{hu2024wavllm}, SALMONN \citep{tang2024salmonn}, MiniCPM-o 2.6 and 4.5 \citep{minicpmo2025}, and Phi-4 Multimodal \citep{abdin2025phi4mm}.
These models differ in backbone, training data, scale, and adaptation, so the main comparison ranks complete systems.

\begin{table*}[t]
\centering
\fontsize{7pt}{8pt}\selectfont
\setlength{\tabcolsep}{1pt}
\begin{tabular}{@{}l*{18}{c}@{}}
\toprule
& \multicolumn{12}{c}{VoiceBench} & \multicolumn{4}{c}{MMAU Test-mini} & \multicolumn{2}{c}{LibriSpeech} \\
\cmidrule(lr){2-13}\cmidrule(lr){14-17}\cmidrule(lr){18-19}
Model & Alp. & Alp.-F & Common & SD-QA & Wild & AdvB & IFEval & OBQA & MMSU & SpMMLU & BBH & Overall & Sound & Music & Speech & Avg & Clean & Other \\
\midrule
\textbf{\model{}} & 82.91 & 79.12 & 75.50 & 41.94 & \textbf{69.10} & 99.62 & 64.30 & \textbf{90.33} & \textbf{72.32} & \textbf{73.42} & \textbf{80.40} & \textbf{75.36} & 65.77 & 60.78 & 60.96 & 62.50 & 2.39 & 5.07 \\
\midrule
Kimi-Audio & 76.00 & 74.60 & 68.40 & 49.52 & 66.20 & \textbf{100.00} & 58.20 & 82.40 & 61.70 & 67.40 & 68.40 & 70.26 & 73.30 & 57.20 & 68.20 & 66.23 & \textbf{1.40} & \textbf{2.59} \\
MiMo-Audio & 75.60 & 74.40 & 62.40 & 33.19 & 63.20 & 98.10 & 68.10 & 75.80 & 58.70 & 56.00 & 71.60 & 67.01 & \textbf{82.30} & \textbf{75.80} & 71.80 & \textbf{76.63} & 3.19 & 6.40 \\
MiniCPM-o 4.5 & \textbf{84.20} & \textbf{81.20} & \textbf{75.60} & \textbf{54.79} & 68.60 & 96.70 & \textbf{78.00} & 83.70 & 69.70 & 40.70 & 70.20 & 73.04 & 77.80 & 71.90 & 73.00 & 74.23 & 1.72 & 3.98 \\
Qwen2.5-Omni & 69.00 & 68.20 & 64.60 & 43.96 & 58.80 & 99.40 & 51.40 & 81.10 & 56.10 & 66.50 & 66.90 & 66.00 & 81.40 & 67.70 & \textbf{75.40} & 74.83 & 1.85 & 3.53 \\
Audio Flamingo 3$^{\dagger}$ & 79.60 & 73.00 & 66.20 & 35.71 & 57.20 & 99.40 & 29.20 & 65.90 & 47.40 & 50.70 & 51.80 & 59.65 & 76.90 & 71.90 & 50.50 & 66.43 & 2.15 & 3.56 \\
MiniCPM-o 2.6 & 81.00 & 74.80 & 60.80 & 49.66 & 56.20 & 98.80 & 42.40 & 62.90 & 43.20 & 34.30 & 60.20 & 60.39 & 78.70 & 68.90 & 70.30 & 72.63 & 1.81 & 4.58 \\
Phi-4 Multimodal & 71.60 & 67.60 & 66.40 & 39.64 & 57.00 & 93.50 & 45.60 & 64.60 & 41.80 & 48.30 & 61.60 & 59.79 & 61.90 & 52.10 & 64.90 & 59.63 & 1.76 & 3.93 \\
Qwen2-Audio & 65.60 & 62.80 & 59.60 & 25.35 & 49.60 & 99.20 & 27.60 & 50.10 & 36.70 & 37.80 & 52.50 & 51.53 & 60.10 & 51.20 & 46.60 & 52.63 & 2.29 & 4.96 \\
WavLLM & 69.20 & 65.80 & 58.40 & 29.56 & 51.40 & 58.70 & 24.30 & 28.10 & 23.30 & 15.10 & 44.40 & 42.57 & 37.20 & 20.40 & 23.40 & 27.00 & 2.08 & 5.01 \\
SALMONN-13B & 26.00 & 26.20 & 31.40 & 15.55 & 27.00 & 3.50 & 17.30 & 23.50 & 23.50 & 25.00 & 46.10 & 24.10 & 49.70 & 36.80 & 33.10 & 39.87 & 2.25 & 5.24 \\
\bottomrule
\end{tabular}
\caption{\textbf{Overall Task Performance.}
We report GPT-5.4 judge scores rescaled to 0--100, task accuracy, refusal rate, and word error rate (WER) on VoiceBench, MMAU Test-mini, and LibriSpeech.
Overall is the equal-weight mean of all eleven VoiceBench tasks, excluding MMAU and ASR.
Bold marks the best result in each column, and lower is better for WER.
Alp., Alp.-F, Common, Wild, AdvB, OBQA, and SpMMLU denote AlpacaEval, AlpacaEval-Full, CommonEval, WildVoice, AdvBench, OpenBookQA, and SpeechMMLU.
$^{\dagger}$Audio Flamingo 3 uses its transcription checkpoint for LibriSpeech and its chat checkpoint for the other benchmarks.}
\label{tab:voicebench-main}
\end{table*}

\subsubsection{Benchmarks}

We evaluate \model{} and the baseline models on three benchmarks:

\smallskip
\noindent\textbullet\enspace\textbf{VoiceBench} \citep{chen2024voicebench} evaluates general speech capabilities, including spoken interaction, instruction following, question answering, reasoning, and refusal behavior.
GPT-5.4 judges response quality on a 1--5 scale for AlpacaEval, AlpacaEval-Full, CommonEval, and WildVoice, which we rescale to 0--100, and answer correctness for SD-QA.
We report all eleven subsets that we evaluate and compute Overall as their equal-weight mean.
The 2B comparisons evaluate each model with both spoken and written inputs.

\smallskip
\noindent\textbullet\enspace\textbf{LibriSpeech test-clean and test-other} \citep{panayotov2015librispeech} evaluate automatic speech recognition.
We report corpus-level word error rate (WER) for each split.

\smallskip
\noindent\textbullet\enspace\textbf{MMAU Test-mini} \citep{sakshi2025mmau} evaluates audio understanding and reasoning across sound, music, and speech.
We report accuracy for each category and the benchmark's average score.

\smallskip
Appendix~\ref{app:baseline-provenance} gives the prompts, decoding and judge settings, and baseline details.

\section{Experimental Results}
\label{sec:results}

\Cref{sec:results-main} compares \model{} with the baselines in Table~\ref{tab:voicebench-main}, \Cref{sec:results-bridge} compares interfaces under a shared Stage 3 recipe, and \Cref{sec:results-topk} varies the support size $K$.
Training details are provided in Appendix~\ref{app:reproducibility} and evaluation protocols in Appendix~\ref{app:baseline-provenance}.

\subsection{Overall Task Performance}
\label{sec:results-main}

Across the eleven VoiceBench tasks, \model{} achieves an Overall score of 75.36, the highest in Table~\ref{tab:voicebench-main}, ahead of MiniCPM-o 4.5 at 73.04 and Kimi-Audio at 70.26.

Its largest margins are on spoken knowledge and reasoning.
\model{} achieves 90.33 on OpenBookQA, 72.32 on MMSU, 73.42 on SpeechMMLU, and 80.40 on BBH, the highest score in each of these columns.
Its BBH score exceeds the next best model, MiMo-Audio at 71.60, by 8.80 points.

On the open-response tasks, \model{} ranks first on WildVoice with 69.10 and second to MiniCPM-o 4.5 on AlpacaEval, AlpacaEval-Full, and CommonEval.
Instruction following and spoken question answering leave the most room, with 64.30 on IFEval and 41.94 on SD-QA against 78.00 and 54.79 for MiniCPM-o 4.5.

\begin{table*}[t]
\centering
\fontsize{7pt}{8pt}\selectfont
\setlength{\tabcolsep}{1pt}
\begin{tabular}{@{}ll*{18}{c}@{}}
\toprule
& & \multicolumn{12}{c}{VoiceBench} & \multicolumn{4}{c}{MMAU Test-mini} & \multicolumn{2}{c}{LibriSpeech} \\
\cmidrule(lr){3-14}\cmidrule(lr){15-18}\cmidrule(lr){19-20}
Interface & Input & Alp. & Alp.-F & Common & SD-QA & Wild & AdvB & IFEval & OBQA & MMSU & SpMMLU & BBH & Overall & Sound & Music & Speech & Avg & Clean & Other \\
\midrule
Base & Text & 57.05 & 60.25 & 57.93 & 23.89 & 57.29 & 99.04 & 64.48 & 69.67 & 56.44 & 43.39 & 76.60 & 60.55 & - & - & - & - & - & - \\
\midrule
MLP & Text & 66.03 & 64.43 & 66.50 & 29.11 & 62.96 & 99.04 & 63.68 & 67.03 & 51.04 & 56.47 & 63.60 & 62.72 & - & - & - & - & - & - \\
 & Speech & 68.54 & 65.91 & 63.30 & 27.75 & 55.56 & 99.04 & 59.73 & 52.97 & 41.28 & 43.63 & 69.10 & 58.80 & 63.66 & 58.38 & 60.06 & 60.70 & 2.10 & 4.60 \\
\midrule
Q-Former & Text & 64.72 & 64.81 & 65.80 & 27.97 & 62.96 & 98.27 & 58.76 & 67.03 & 51.53 & 59.90 & 56.20 & 61.63 & - & - & - & - & - & - \\
 & Speech & 70.15 & 62.39 & 64.30 & 26.94 & 51.56 & 97.12 & 52.59 & 43.08 & 32.14 & 33.43 & 62.40 & 54.19 & 67.57 & 57.49 & 59.16 & 61.40 & 2.49 & 5.52 \\
\midrule
\textbf{\cgate{} (ours)} & Text & 66.73 & 65.28 & 66.30 & 29.84 & 63.12 & 99.23 & 63.38 & 65.93 & 49.32 & 57.88 & 68.20 & 63.20 & - & - & - & - & - & - \\
 & Speech & 69.65 & 66.45 & 63.50 & 27.12 & 55.66 & 100.00 & 61.20 & 59.78 & 43.92 & 46.52 & 70.00 & 60.35 & 63.66 & 55.39 & 59.16 & 59.40 & 2.46 & 5.54 \\
\bottomrule
\end{tabular}
\caption{\textbf{Comparison of Interfaces.}
We train \cgate{}, MLP, and Q-Former interfaces with Qwen3.5-2B under a shared Stage 3 recipe and evaluate each model with written (Text) and spoken (Speech) inputs.
Base is the original Qwen3.5-2B decoder before adaptation.
Scores are on a 0--100 scale, with higher values better except for WER.
Overall averages the eleven VoiceBench tasks.
Dashes mark tasks that are not evaluated for that input.}
\label{tab:text-bridge-ablation}
\end{table*}

On MMAU Test-mini, \model{} achieves an average accuracy of 62.50\%, with 65.77\% on sound, 60.78\% on music, and 60.96\% on speech, against 76.63\% for MiMo-Audio.
On LibriSpeech, \model{} achieves WER of 2.39\% on test-clean and 5.07\% on test-other, lower than MiMo-Audio at 3.19\% and 6.40\% and higher than the other baselines, which Kimi-Audio leads at 1.40\% and 2.59\%.
\model{} also achieves lower WER than the 2B \cgate{} model in Table~\ref{tab:text-bridge-ablation}, by 0.08 percentage points on test-clean and 0.47 on test-other (Appendix~\ref{app:recognition-comparison}).
Kimi-Audio and MiMo-Audio report audio training scales more than an order of magnitude above the adaptation data of \model{} (Appendix~\ref{app:training-scale}).

\subsection{Comparison of Interfaces}
\label{sec:results-bridge}

We compare \cgate{}, MLP, and Q-Former interfaces of matched size with the same encoder and Qwen3.5-2B decoder.
The three models share the Stage 3 training data, objective, effective batch size, learning rates, and schedule, after the same numbers of Stage 1 and Stage 2 updates (Appendix~\ref{app:reproducibility}).
We evaluate each model with spoken inputs and, to measure what the decoder retains, with written versions of the same eleven VoiceBench tasks, which bypass the encoder and interface.
The Base row of Table~\ref{tab:text-bridge-ablation} reports the original Qwen3.5-2B decoder on these written tasks.

\noindent\textbf{Spoken-Language Performance.}
\cgate{} achieves a VoiceBench Overall of 60.35, compared with 58.80 for MLP and 54.19 for Q-Former.
Its largest advantage over MLP is on OpenBookQA, at 59.78 versus 52.97, with further gains on IFEval, MMSU, SpeechMMLU, and BBH.
MLP scores higher on SD-QA, while Q-Former scores higher on AlpacaEval and CommonEval.

\noindent\textbf{Text Performance After Adaptation.}
With written inputs, \cgate{}, MLP, and Q-Former obtain Overall scores of 63.20, 62.72, and 61.63, compared with 60.55 for the original decoder.
All three adapted decoders gain on the open-response subsets, SD-QA, and SpeechMMLU and lose on IFEval, OpenBookQA, MMSU, and BBH.
The largest loss is on BBH, from 76.60 to 68.20 for \cgate{}, 63.60 for MLP, and 56.20 for Q-Former.
\cgate{} also has the smallest gap between written and spoken inputs, at 2.86 points against 3.92 for MLP and 7.44 for Q-Former.

\noindent\textbf{Recognition and Audio Understanding.}
MLP achieves the lowest LibriSpeech WER, at 2.10\% on test-clean and 4.60\% on test-other, while Q-Former has the highest MMAU accuracy at 61.40\%.
Relative to \cgate{}, MLP reduces WER by 0.36 and 0.94 percentage points, chiefly through fewer substitutions, while \cgate{} gains most of its VoiceBench advantage on OpenBookQA, MMSU, and SpeechMMLU (Appendix~\ref{app:recognition-comparison}).
The additional tasks in Table~\ref{tab:app-18k-audio} follow the same pattern, with \cgate{} ahead on spoken interaction and question answering and MLP on transcription.
Each interface therefore leads on a different task family.

\subsection{Comparison of \texorpdfstring{Top-$K$}{Top-K} Configurations}
\label{sec:results-topk}

\begin{figure*}[t]
\centering
\includegraphics[width=\textwidth]{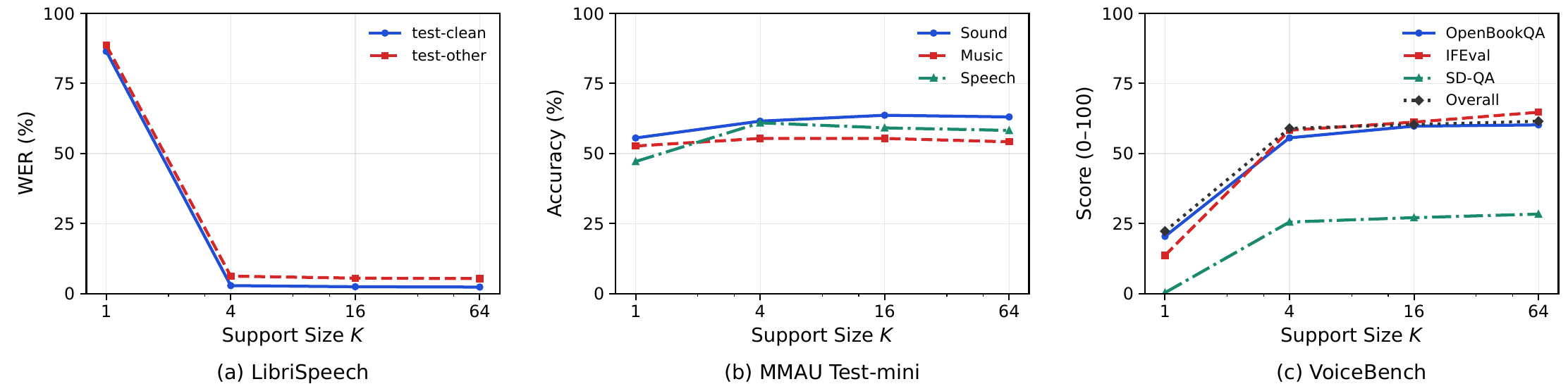}
\caption{\textbf{Comparison of Top-$K$ Configurations.}
We train \cgate{} with support sizes $K \in \{1, 4, 16, 64\}$ on Qwen3.5-2B.
(a) LibriSpeech test-clean and test-other WER in percent.
(b) MMAU Test-mini accuracy for sound, music, and speech.
(c) Speech-input OpenBookQA, IFEval, SD-QA, and Overall scores on a 0--100 scale, where Overall averages the eleven VoiceBench tasks of Table~\ref{tab:text-bridge-ablation}.
The $K=1$ run uses a straight-through gradient (Appendix~\ref{app:efficiency}).}
\label{fig:topk-ablation}
\end{figure*}

We vary the support size $K\in\{1,4,16,64\}$ in Stage 3, starting all four runs from the same Stage 2 checkpoint (Figure~\ref{fig:topk-ablation}, Appendix~\ref{app:efficiency}).
With a single embedding per position, WER exceeds 86\% on both LibriSpeech splits and the VoiceBench Overall is 22.30, so mixing several embeddings is essential.
From $K=4$ to $K=64$, WER decreases and the OpenBookQA, IFEval, SD-QA, and Overall scores increase at every step.
MMAU does not follow this trend.
From $K=4$ to $K=16$, sound accuracy rises, music accuracy is unchanged, and speech accuracy falls, and at $K=64$ all three categories are lower than at $K=16$, with an average of 58.50\% against 59.40\%.
The setting used in all other experiments, $K=16$, has the highest MMAU average, and $K=64$ further improves WER and VoiceBench Overall while mixing four times as many embeddings per position.

\section{Analysis}
\label{sec:analysis}

\begin{figure*}[t]
\centering
\includegraphics[width=\textwidth]{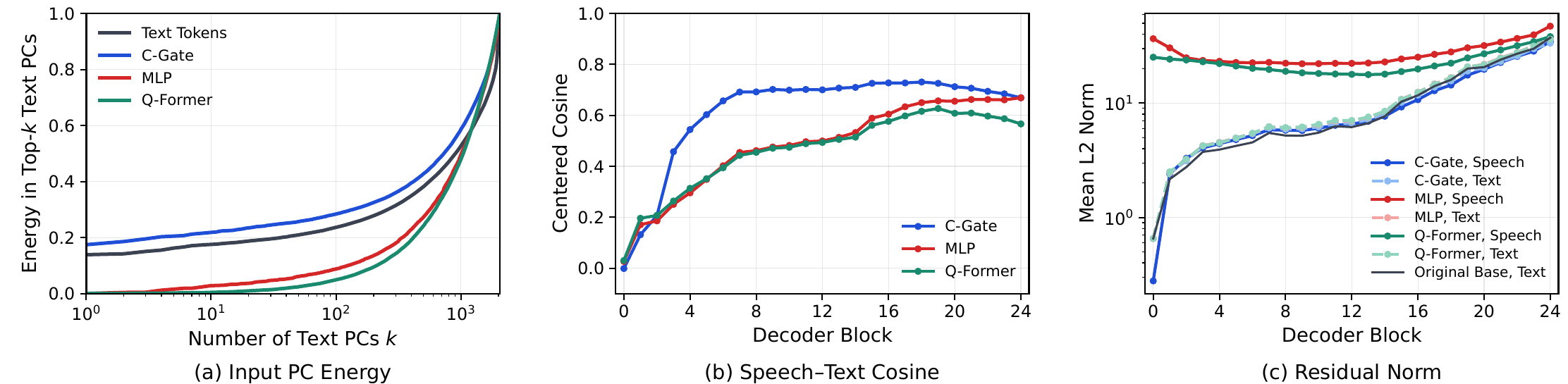}
\caption{\textbf{Speech Representations Relative to Text Representations.}
We compare 2B \cgate{}, MLP, and Q-Former models (Appendix~\ref{app:representation-measurements}) on 200 spoken questions and their written counterparts.
(a) Cumulative fraction of input variation, or energy, captured by the leading principal directions of the text embedding table.
(b) Cosine similarity between the spoken and written versions of each question after each decoder block, where block 0 is the decoder input.
(c) Mean L2 norm of speech and text representations after each decoder block, where Original Base is the original Qwen3.5-2B decoder.}
\label{fig:interface-geometry}
\end{figure*}

We compare where the three interfaces place speech in the decoder's representation space.
The measurements use a separate set of 2B \cgate{}, MLP, and Q-Former models trained for 6,000 Stage 3 updates, and 200 questions available in both spoken and written form (Appendix~\ref{app:representation-measurements}).

\noindent\textbf{Input Geometry.}
\begin{figure}[t]
\centering
\includegraphics[width=\columnwidth]{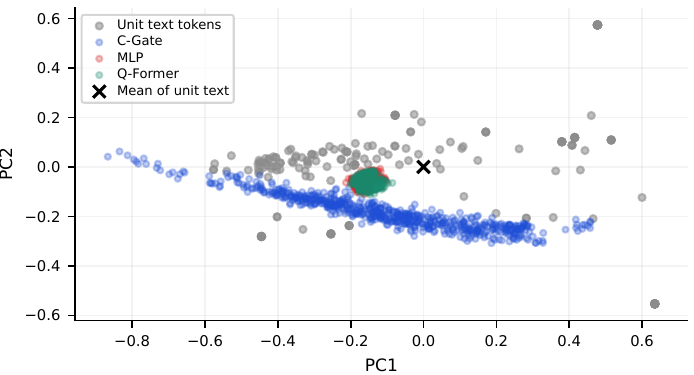}
\caption{\textbf{PCA of Normalized Decoder Inputs.}
After unit L2 normalization, we project the decoder inputs of the three models in Figure~\ref{fig:interface-geometry} onto the first two principal components of the written tokens from 20 questions.}
\label{fig:input-pca-overlay}
\end{figure}

\cgate{} concentrates more of its input variation in the leading principal directions of the text embedding table than MLP or Q-Former (Figure~\ref{fig:interface-geometry}a).
The first 100 text directions hold 28.4\% of the variation of \cgate{} inputs and 23.7\% for written-question tokens, against 8.8\% for MLP and 4.9\% for Q-Former.
Figure~\ref{fig:input-pca-overlay} projects unit-normalized inputs for 20 of the questions onto the first two principal components of their written tokens.
\cgate{} inputs spread across the region occupied by text tokens, while MLP and Q-Former inputs fall in a small cluster.

\noindent\textbf{Alignment Across Decoder Blocks.}
Figure~\ref{fig:interface-geometry}b reports the cosine similarity between the spoken and written versions of each question after subtracting the mean over the other questions, which removes what all questions share.
\cgate{} reaches 0.69 by block 7 and stays between 0.69 and 0.73 through block 19, while MLP and Q-Former are at 0.45 and 0.44 at block 7.
MLP reaches the level of \cgate{} only at the final block, at 0.67, and Q-Former peaks at 0.63.

\noindent\textbf{Representation Scale.}
\cgate{} speech representations follow the norm of the written-question representations from the first decoder block onward (Figure~\ref{fig:interface-geometry}c).
MLP and Q-Former inputs enter the decoder with norms of 36.7 and 25.2, against 0.65 for text embeddings, a scale that text representations reach only in the last few decoder blocks.

\cgate{} therefore places speech inputs in the leading text directions, at text scale, and aligned with the paired written question from the early decoder blocks.

\section{Conclusion}
\label{sec:discussion}

We presented \model{}, a general-purpose speech and audio language model that represents audio as continuous convex combinations of frozen text embeddings.
\model{} leads the compared models on the eleven-task VoiceBench aggregate, with its largest margins on spoken knowledge and reasoning, while the strongest baselines are ahead on MMAU and transcription.
In a controlled 2B comparison, \cgate{} has the highest aggregate spoken-language performance among the three interfaces, text-task performance close to MLP, and the smallest gap between written and spoken inputs.
MLP performs better on transcription, and Q-Former has the highest MMAU accuracy.
These results support convex combinations of text embeddings as a viable interface for general-purpose speech and audio language models.

\section*{Limitations}
\label{sec:limitations}

Our main comparison spans systems that differ in scale, training data, and adaptation recipe, so it compares complete systems rather than interfaces in isolation.
The interface comparison trains one model per interface at the 2B scale, and the representation analysis uses a separate set of 2B models trained with a shorter schedule and a per-task loss average (Appendix~\ref{app:representation-models}).

\section*{Ethical Considerations}
\label{sec:ethics}

Speech recordings may contain identifying and sensitive personal information, and their use should respect speakers' consent and privacy.
Biases in training data and synthetic speech may lead to uneven performance across accents and speaker groups, which we have not systematically evaluated.
\model{} may also produce incorrect or harmful responses.

\bibliography{references}

\appendix
\section{Implementation and Training Details}
\label{app:reproducibility}

This appendix documents the training of the 9B model (Table~\ref{tab:voicebench-main}), the 2B interface and Top-$K$ comparisons (Table~\ref{tab:text-bridge-ablation} and Figure~\ref{fig:topk-ablation}), and the models used for the representation analysis.
Appendix~\ref{app:data-audit} describes the training data, Appendix~\ref{app:baseline-provenance} specifies evaluation, and Appendix~\ref{app:bridge-matrix} defines the representation measurements.

\subsection{Architecture and Parameter Adaptation}
\label{app:parameter-adaptation}

Section~\ref{sec:experimental-setup} gives the architecture and the three training stages, and this subsection adds the remaining implementation details.
Routing scores are clipped to $[-30,30]$.
The router scores the vocabulary in chunks of 2,048 rows and merges the highest-scoring entries across chunks, which yields the same support as scoring all rows at once with less memory.
In Stage 1 the encoder and decoder are frozen, and gradients pass through the decoder to the interface.
Stage 2 additionally trains the encoder while keeping the decoder frozen.

\subsection{Main 9B Training}
\label{app:main-training}

Stages 1 and 2 each run for 7,000 updates with an effective batch size of 2,048.
We evaluate Stage 3 at 7,500 updates with an effective batch size of 512 and an optimizer and learning-rate schedule initialized at the start of the stage.
After Stage 3 update 5,000, we increase the sampling probability of transcription data from 20\% to 35\% and of ASR-derived instructions from 10\% to 15\%, together with changes to the other audio task proportions.

In Stage 3, audio examples that have a written form are trained with the distillation loss of Equation~\ref{eq:distill-loss}, using a frozen copy of the original Qwen3.5-9B as the text teacher and a distillation temperature of 1.0.
The remaining audio examples and text replay use cross-entropy.
Stage 3 normalizes the loss by the total valid response-token count across data-parallel ranks and gradient-accumulation microsteps in each optimizer update.

The Stage 3 learning rate follows a warmup--stable--decay schedule defined over 12,000 updates, with 200 warmup updates and a constant learning rate until update 4,000.
Training uses 16 H100 GPUs.
Peak learning rates are $10^{-4}$ for the interface, $10^{-5}$ for the encoder, and $2\times10^{-5}$ for the decoder.
The weight decay is 0.01 and the minimum learning-rate ratio is 0.1.

\subsection{2B Interfaces and Training Objective}
\label{app:representation-models}

We compare \cgate{} with an MLP and a Transformer-based Q-Former using the same acoustic encoder and 2B decoder.
Their interface parameter counts differ by less than 0.17\%.
Table~\ref{tab:app-2b-training} summarizes the architecture and Stage 3 training settings, which the three interfaces share together with the training data.
Each interface receives 5,000 cumulative Stage 1 updates, 7,000 Stage 2 updates, and 18,000 Stage 3 updates.
Stage 1 comprises separately initialized optimizer segments of 2,000 and 3,000 updates for \cgate{}, 2,500 and 2,500 for MLP, and 2,250, 250, and 2,500 for Q-Former, with model weights carried between segments.
These segments share an effective batch size of 2,048, as does Stage 2.
The Stage 3 learning rate follows a cosine schedule defined over 26,000 updates, and we compare checkpoints at update 18,000.
Stage 3 uses teacher-forced cross-entropy with the same global response-token normalization described above.
Stage 3 uses eight H100 GPUs.

\begin{table*}[t]
\centering
\fontsize{9pt}{10.5pt}\selectfont
\setlength{\tabcolsep}{4pt}
\renewcommand{\arraystretch}{1.12}
\begin{tabular}{@{}>{\raggedright\arraybackslash}p{.32\textwidth} >{\raggedright\arraybackslash}p{.20\textwidth} >{\raggedright\arraybackslash}p{.20\textwidth} >{\raggedright\arraybackslash}p{.20\textwidth}@{}}
\toprule
\textbf{Setting} & \textbf{MLP} & \textbf{Q-Former} & \textbf{\cgate{}} \\
\midrule
Acoustic encoder & Whisper-large-v3 & Whisper-large-v3 & Whisper-large-v3 \\
Decoder & Qwen3.5-2B & Qwen3.5-2B & Qwen3.5-2B \\
Interface parameters & 1,706,494 & 1,708,268 & 1,705,473 \\
Temporal pooling stride & 4 & 4 & 4 \\
Routing dimension & N/A & N/A & 512 \\
Support size $K$ & N/A & N/A & 16 \\
Initial routing temperature & N/A & N/A & 1.0 \\
Mixture temperature & N/A & N/A & 0.7 \\
Training dropout & 0.1 & 0.1 & 0.1 \\
Input embeddings / output head & Frozen & Frozen & Frozen \\
Interface learning rate & $3\times10^{-5}$ & $3\times10^{-5}$ & $3\times10^{-5}$ \\
Encoder learning rate & $3\times10^{-6}$ & $3\times10^{-6}$ & $3\times10^{-6}$ \\
Decoder learning rate & $10^{-5}$ & $10^{-5}$ & $10^{-5}$ \\
Weight decay & 0.01 & 0.01 & 0.01 \\
Effective batch size & 256 & 256 & 256 \\
Warm-up updates & 200 & 200 & 200 \\
Minimum learning-rate ratio & 0.05 & 0.05 & 0.05 \\
\bottomrule
\end{tabular}
\caption{Architecture and training settings for the 2B interface comparison.}
\label{tab:app-2b-training}
\end{table*}

The MLP applies two affine transformations, from 1,280 to 510 and then to 2,048 dimensions, with GELU between them and LayerNorm at the input and output.
The Q-Former uses 64 learned queries of width 280 and one eight-head cross-attention block over pooled acoustic features, followed by a feed-forward sublayer of width 420.
Its attention and feed-forward sublayers use pre-normalization and residual connections.
A final LayerNorm followed by a linear projection maps the queries into the 2,048-dimensional decoder input space.
Both interfaces use dropout of 0.1.
Q-Former produces 64 decoder-input positions, whereas MLP and \cgate{} preserve the length of the pooled acoustic sequence.

\noindent\textbf{Models for the Representation Analysis.}
The representation analysis of Section~\ref{sec:analysis} uses a separate set of 2B \cgate{}, MLP, and Q-Former models, trained for 6,000 Stage 3 updates with an effective batch size of 2,048.
Their loss averages the cross-entropy within each task and then combines the task means, whereas the models of Table~\ref{tab:text-bridge-ablation} average over all response tokens.

\subsection{Recognition Differences}
\label{app:recognition-comparison}

We evaluate the 9B model and the 2B interfaces on LibriSpeech with the verbatim-repetition instruction of their Stage 3 raw-ASR training (Appendix~\ref{app:inference-protocol}).
The 9B model achieves lower WER than the 2B \cgate{} model, by 0.08 percentage points on test-clean and 0.47 on test-other, with paired speaker-bootstrap 95\% intervals of [\textminus 0.28,~0.16] and [\textminus 0.67,~\textminus 0.28] for the 9B-minus-2B difference.
The two models also differ in Stage 3 training, where the 9B model learns raw-ASR examples through a frozen text teacher and the 2B models use cross-entropy (Appendices~\ref{app:main-training} and~\ref{app:representation-models}).

For the two 2B interfaces, \cgate{} makes 192 more word edits than MLP on test-clean and 497 more on test-other.
Substitutions account for 559 of these 689 additional edits (81.1\%).
The WER differences are 0.36 and 0.94 percentage points, with paired speaker-bootstrap 95\% intervals of [0.24, 0.48] and [0.73, 1.19], using 10,000 resamples of the 40 clean and 33 other speakers.
The complementary VoiceBench advantage is concentrated in OpenBookQA, MMSU, and SpeechMMLU, which contribute 1.12 of the 1.54 aggregate points separating \cgate{} and MLP.
Their respective task differences are 6.81, 2.64, and 2.89 points, with paired-item 95\% intervals of [1.98, 11.65], [0.68, 4.55], and [1.71, 4.07] over 5,000 resamples.

\section{Training Data and Preprocessing}
\label{app:data-audit}

\subsection{Data Composition}

Table~\ref{tab:app-2b-data} groups the 45 training sources of the 2B comparison into four families with their sampling probabilities.
Recognition data are the public corpora listed in Section~\ref{sec:experimental-setup}.
The ASR-derived instruction family consists of spoken instruction examples derived from transcription data.
The third family covers audio captioning, sound-event understanding, music questions, emotion recognition, and spoken instruction and reasoning, where the instruction and reasoning sources use synthesized speech.
Emotion supervision comes from public emotion-recognition corpora.
Text replay consists of three text-only sources for dialogue, question answering, and instruction following.

\begin{table*}[t]
\centering
\tablebodyfont
\setlength{\tabcolsep}{4pt}
\renewcommand{\arraystretch}{1.12}
\begin{tabular*}{\textwidth}{@{}p{.20\textwidth}@{\extracolsep{\fill}}p{.49\textwidth}cr@{}}
\toprule
\textbf{Task Family} & \textbf{Dataset or Source Group} & \textbf{Input} & \textbf{Probability} \\
\midrule
Recognition & MLS English, GigaSpeech, LibriSpeech, Emilia, Common Voice English, FLEURS English & Audio & 35.0\% \\
ASR-derived instruction & Transcription-derived spoken instruction sources & Audio & 15.0\% \\
Interaction and audio understanding & Public audio, synthetic speech, and in-house data & Audio & 35.0\% \\
Text replay & General dialogue, question answering, instruction following & Text & 15.0\% \\
\textbf{Total} & & & \textbf{100.0\%} \\
\bottomrule
\end{tabular*}
\caption{Source-sampling probabilities for the 2B interface comparison.}
\label{tab:app-2b-data}
\end{table*}

The main 9B model draws on the same task families.
The 2B comparison uses the final Stage 3 sampling proportions of the 9B model, the ones in effect after the weight of transcription data is raised, throughout its own Stage 3.

\subsection{Training Data Scale}
\label{app:training-scale}

Table~\ref{tab:app-training-scale} lists the audio training hours that each main-table baseline reports, which cover different training phases across systems, together with the audio used to adapt \model{}.
Adaptation presents 14.3M audio examples in each of Stages 1 and 2 and 3.8M examples in Stage 3, some of which are text replay.
Counting every example as audio at the 30-second input cap bounds the adaptation audio at 0.271M hours.
The pool from which these examples are sampled contains 57.2M audio annotations and 0.44M text examples.
These figures cover adaptation and exclude the pretraining of the Whisper-large-v3 encoder, which used 1M hours of weakly labeled and 4M hours of pseudo-labeled audio~\citep{openai2023whisperlargev3}, and the pretraining of the Qwen3.5 decoder.

\begin{table*}[t]
\centering
\fontsize{9pt}{10.5pt}\selectfont
\setlength{\tabcolsep}{4pt}
\renewcommand{\arraystretch}{1.15}
\begin{tabular*}{\textwidth}{@{}l@{\extracolsep{\fill}}r@{}}
\toprule
\textbf{Model} & \textbf{Audio Hours} \\
\midrule
Kimi-Audio~\citep{kimiaudio2025} & $\approx$13M \\
MiMo-Audio~\citep{mimoaudio2025} & $>$100M \\
MiniCPM-o 4.5~\citep{cui2026minicpmo45} & Millions \\
Qwen2.5-Omni~\citep{xu2025qwen25omni} & -- \\
Audio Flamingo 3~\citep{nvidia2025audioflamingo3data} & -- \\
MiniCPM-o 2.6~\citep{openbmb2025minicpmo26} & -- \\
Phi-4 Multimodal~\citep{abdin2025phi4mm} & $\approx$2M \\
Qwen2-Audio~\citep{chu2024qwen2audio} & 0.52M \\
WavLLM~\citep{hu2024wavllm} & 7.7K / 2.9K \\
SALMONN-13B~\citep{tang2024salmonn} & $\approx$4.4K \\
\midrule
\model{} (ours, 9B) & $\leq$0.271M \\
\bottomrule
\end{tabular*}
\caption{Audio training hours for the baseline models and our model, in millions (M) or thousands (K).}
\label{tab:app-training-scale}
\end{table*}

Kimi-Audio and MiMo-Audio report audio training scales more than an order of magnitude above this bound, and \model{} obtains the higher VoiceBench aggregate in Table~\ref{tab:voicebench-main}.
In Table~\ref{tab:app-training-scale}, dashes mark systems that report no total, the MiniCPM-o 4.5 report gives its scale only as millions of hours, and the two WavLLM values are its single-task and multitask training sets.

\subsection{Preprocessing and Held-Out Partitions}

All training inputs are limited to 30 seconds of audio, and temporal mean pooling with stride four is applied to the acoustic encoder outputs.
The 2B FLEURS source combines the English training and validation partitions and reserves the test partition for evaluation.

\raggedbottom
\section{Evaluation Protocols and Baseline Details}
\label{app:baseline-provenance}

\subsection{Benchmarks and Score Aggregation}

We evaluate spoken-language capabilities with eleven VoiceBench subsets in the main 9B comparison, audio understanding with MMAU Test-mini, and transcription with LibriSpeech test-clean and test-other.
Table~\ref{tab:app-eval-protocol} lists the evaluation subset sizes, generation budgets, and reported metrics.

\begin{table*}[t]
\centering
\fontsize{9pt}{10.5pt}\selectfont
\setlength{\tabcolsep}{4pt}
\renewcommand{\arraystretch}{1.12}
\begin{tabular}{@{}>{\raggedright\arraybackslash}p{.26\textwidth} >{\centering\arraybackslash}p{.09\textwidth} >{\centering\arraybackslash}p{.10\textwidth} >{\raggedright\arraybackslash}p{.47\textwidth}@{}}
\toprule
\textbf{Benchmark / Subset} & \textbf{Items} & \textbf{Max. Tokens} & \textbf{Reported Metric} \\
\midrule
VoiceBench: AlpacaEval & 199 & 2,048 & Mean quality score (1--5) $\times$20 \\
VoiceBench: AlpacaEval-Full & 636 & 2,048 & Mean quality score (1--5) $\times$20 \\
VoiceBench: CommonEval & 200 & 2,048 & Mean quality score (1--5) $\times$20 \\
VoiceBench: SD-QA & 6,083 & 2,048 & GPT-5.4 judged correctness (\%) \\
VoiceBench: WildVoice & 1,000 & 2,048 & Mean quality score (1--5) $\times$20 \\
VoiceBench: AdvBench & 520 & 2,048 & Deterministic refusal rate (\%) \\
VoiceBench: IFEval & 345 & 2,048 & Mean of four strict/loose scores (\%) \\
VoiceBench: OpenBookQA & 455 & 2,048 & Parsed-choice accuracy (\%) \\
VoiceBench: MMSU & 3,074 & 2,048 & Parsed-choice accuracy (\%) \\
VoiceBench: SpeechMMLU & 8,549 & 2,048 & Parsed-choice accuracy (\%) \\
VoiceBench: BBH & 1,000 & 2,048 & Category-aware answer accuracy (\%) \\
MMAU Test-mini & 1,000 & 256 & Parsed-choice accuracy by category and overall (\%) \\
LibriSpeech test-clean & 2,620 & 256 & Corpus word error rate (\%) \\
LibriSpeech test-other & 2,939 & 256 & Corpus word error rate (\%) \\
\bottomrule
\end{tabular}
\caption{Overview of evaluation subsets.}
\label{tab:app-eval-protocol}
\end{table*}

We refer to AlpacaEval, AlpacaEval-Full, CommonEval, and WildVoice as the open-response subsets.
For these subsets, GPT-5.4 assigns response-quality scores from 1 to 5, and we multiply the mean score by 20.
SD-QA uses GPT-5.4 to judge answer correctness, and AdvBench measures the refusal rate by matching refusal strings.
IFEval, OpenBookQA, MMSU, SpeechMMLU, and BBH use their standard automatic scoring.
VoiceBench Overall is the equal-weight mean of the eleven subset scores on the 100-point scale, in the main comparison and in the 2B comparisons.

For IFEval, we report the arithmetic mean of strict prompt-level, strict instruction-level, loose prompt-level, and loose instruction-level accuracy.
For \model{}, these are 59.13, 69.08, 59.71, and 69.28\%.

MMAU Test-mini contains 1,000 questions (333 sound, 334 music, and 333 speech), scored by the parsed answer choice and reported separately for the three categories.

For LibriSpeech, corpus-level WER divides the total number of substitutions, deletions, and insertions by the total number of reference words.
Baseline ASR results use greedy decoding with each model's own transcription prompt.

\subsection{Prompts, Decoding, and Judge Configuration}
\label{app:inference-protocol}

We use greedy decoding for \model{}.
The open-response subsets, SD-QA, AdvBench, and IFEval present the spoken input without an additional written task instruction.
OpenBookQA, MMSU, SpeechMMLU, and BBH use fixed task instructions.
Output budgets are listed in Table~\ref{tab:app-eval-protocol}.

Each example receives one GPT-5.4 judgment with a 1,024-token completion budget.
The main-model evaluation uses the judge's default sampling settings, and the 2B comparison uses temperature 0.5, top-$p$ 0.95, and random seed 0.
All three interfaces use the same fixed task prompts.
VoiceBench MMSU contains 3,074 examples and is distinct from the standalone 5,000-example MMSU benchmark.

ASR uses the instruction ``Repeat the sentence you hear exactly, word for word, with no other text.'' for both the 9B model and the 2B comparison, with Whisper English text normalization.
This is the raw-ASR instruction of their Stage 3 training.

\subsection{Additional 2B Speech Tasks}

Table~\ref{tab:app-18k-audio} reports eight additional speech-input tasks for the 2B interface comparison.
\cgate{} scores highest on MTBench~\citep{zheng2023mtbench} and all five OpenAudioBench tasks~\citep{baichuan2025omni15}.
MLP achieves the lowest WER on Common Voice and FLEURS.
OpenAudioBench ReasoningQA is scored by matching the response against accepted answer aliases, and the other OpenAudioBench tasks are judged by GPT-5.4.

\begin{table*}[t]
\centering
\tablebodyfont
\setlength{\tabcolsep}{4pt}
\begin{tabular*}{\textwidth}{@{\extracolsep{\fill}}llrccc@{}}
\toprule
Task & Metric & $N$ & \cgate{} & MLP & Q-Former \\
\midrule
MTBench & GPT 1--5 & 46 & 2.41 & 2.26 & 1.91 \\
OAB AlpacaEval & GPT 1--10 & 199 & 4.85 & 4.66 & 4.61 \\
OAB LlamaQuestions & Correct & 300 & 63.67 & 62.67 & 58.67 \\
OAB TriviaQA & GPT 0--10 & 1,000 & 2.81 & 2.73 & 2.36 \\
OAB WebQuestions & GPT 0--10 & 1,000 & 4.15 & 4.10 & 3.96 \\
OAB ReasoningQA & Alias accuracy & 202 & 6.93 & 4.95 & 4.46 \\
Common Voice 15 en & WER $\downarrow$ & 16,386 & 11.31 & 9.26 & 9.51 \\
FLEURS en-US & WER $\downarrow$ & 647 & 5.22 & 4.57 & 4.72 \\
\bottomrule
\end{tabular*}
\caption{Additional speech-input tasks for the 2B interface comparison of Table~\ref{tab:text-bridge-ablation}.
All metrics are percentages except the GPT score scales given in the Metric column.
Higher is better except for WER.
OAB denotes OpenAudioBench.}
\label{tab:app-18k-audio}
\end{table*}

\subsection{Text and Speech Input Conditions}
\label{app:text-speech-conditions}

The 2B interface comparison evaluates the adapted decoder with both text and speech inputs.
Text input bypasses the acoustic encoder and interface and measures the decoder's task performance after adaptation.
Speech input additionally requires the encoder and interface to provide the information needed for the response.
The Base row reports the original Qwen3.5-2B decoder before adaptation on the same eleven text tasks.

\subsection{Baseline Models}

We run each baseline with its released checkpoint and inference settings and score the open-response subsets with GPT-5.4.
Audio Flamingo 3 uses its transcription checkpoint for ASR and its chat checkpoint for the other benchmarks.

\subsection{Failure Handling}

The quality-score parser extracts a score from 1 to 5, and the SD-QA parser extracts a Yes or No correctness decision.
An empty response receives a score of zero or a No decision and is kept in the average.
For AdvBench, an empty response counts as a non-refusal.
Examples rejected by the judge's content filter, at most eight per task, are excluded from the score.

\flushbottom
\section{Representation Diagnostics}
\label{app:bridge-matrix}
\label{app:text-evaluation}

\subsection{Paired Questions and Hidden-State Extraction}
\label{app:representation-measurements}

Figures~\ref{fig:interface-geometry} and~\ref{fig:input-pca-overlay} are measured on the 2B \cgate{}, MLP, and Q-Former models described in Appendix~\ref{app:representation-models}.
We select 200 OpenAudioBench LlamaQuestions examples with audio longer than one second and no longer than 15 seconds and a nonempty reference answer of at most 200 characters.
We shuffle the eligible examples with a fixed seed and retain the first 200 distinct written questions.
The first 20 questions form the PCA visualization subset.

We run each model on the spoken question and on its written counterpart without generating or supplying answer tokens.
The spoken input contains the audio and no written question.
For each input, we extract the audio or question span at the decoder input and after each of the 24 decoder blocks, before the final normalization layer.
We exclude the surrounding prompt tokens from span measurements.
The decoder states have 2,048 dimensions.
MLP and \cgate{} produce one position per pooled audio frame, while Q-Former produces 64 query positions.

\subsection{Energy in Text Principal Directions}

For Figure~\ref{fig:interface-geometry}a, we fit PCA to the 248,077 embedding rows covered by the tokenizer, excluding extra padding rows in the 248,320-row embedding table.
We order the 2,048 principal directions by descending embedding variance.
For each interface, we concatenate the decoder-input vectors across the 200 questions and subtract that interface's mean vector.
We apply the same procedure to the written-question token embeddings used as the reference curve.
If $X_c$ is this centered matrix and $v_j$ is text principal direction $j$, the cumulative energy is
\begin{equation}
  R_X(k)=\frac{\sum_{j=1}^{k}\lVert X_c v_j\rVert_2^2}
  {\lVert X_c\rVert_F^2}.
\end{equation}
Each curve measures directional variation within a representation set and is unchanged by a common rescaling of that set.
Centering each set separately removes its mean displacement from the text embeddings.

\begin{table*}[t]
\centering
\fontsize{8pt}{9pt}\selectfont
\setlength{\tabcolsep}{1.3pt}
\begin{tabular}{@{}l*{18}{c}@{}}
\toprule
& \multicolumn{12}{c}{VoiceBench} & \multicolumn{4}{c}{MMAU Test-mini} & \multicolumn{2}{c}{LibriSpeech} \\
\cmidrule(lr){2-13}\cmidrule(lr){14-17}\cmidrule(lr){18-19}
$K$ & Alp. & Alp.-F & Common & SD-QA & Wild & AdvB & IFEval & OBQA & MMSU & SpMMLU & BBH & Overall & Sound & Music & Speech & Avg & Clean & Other \\
\midrule
1 & 22.11 & 21.07 & 20.60 & 0.35 & 20.06 & 24.42 & 13.67 & 20.44 & 24.72 & 25.34 & 52.50 & 22.30 & 55.56 & 52.69 & 47.15 & 51.80 & 86.48 & 88.65 \\
4 & 70.75 & 65.19 & 62.70 & 25.56 & 53.24 & 99.42 & 58.30 & 55.60 & 43.36 & 43.17 & 71.60 & 58.99 & 61.56 & 55.39 & 60.96 & 59.30 & 2.87 & 6.25 \\
16 & 69.65 & 66.45 & 63.50 & 27.12 & 55.66 & 100.00 & 61.20 & 59.78 & 43.92 & 46.52 & 70.00 & 60.35 & 63.66 & 55.39 & 59.16 & 59.40 & 2.46 & 5.54 \\
64 & 71.16 & 66.92 & 62.30 & 28.39 & 57.08 & 99.81 & 64.74 & 60.22 & 46.94 & 47.61 & 71.70 & 61.53 & 63.06 & 54.19 & 58.26 & 58.50 & 2.35 & 5.37 \\
\bottomrule
\end{tabular}
\caption{Speech-input results for the four Top-$K$ configurations of Figure~\ref{fig:topk-ablation}, in the layout of Table~\ref{tab:text-bridge-ablation}.
Scores are on a 0--100 scale, with higher values better except for WER.
Overall averages the eleven VoiceBench tasks.}
\label{tab:app-topk}
\end{table*}

\subsection{Paired Alignment and Representation Norms}

For Figure~\ref{fig:interface-geometry}b, we first average the positions within each audio or written-question span.
For each modality and decoder block, we then subtract the mean of the other 199 questions from each question's pooled vector, which removes the component shared by all questions.
We compute cosine similarity between the resulting speech and text vectors for the same question and average over the 200 pairs, where the text vector comes from the same adapted model run on the written question.
Across 200 random re-pairings of speech and text, 95\% of the similarities lie within 0.03 of zero at every block for all three models.

For Figure~\ref{fig:interface-geometry}c, we compute the L2 norm at each span position, average within each question, and then average across questions.
This gives each question equal weight despite differences in audio duration or token count.
The original Qwen3.5-2B decoder, run on the written questions, supplies an additional norm reference.

\subsection{Normalized Input PCA Visualization}

Figure~\ref{fig:input-pca-overlay} uses a separate PCA basis fitted to the question tokens from the 20-question subset.
We normalize every input vector to unit L2 norm before centering and fitting this basis.
For each question, we select uniformly spaced audio positions with the same count across interfaces, capped at 32 and bounded by the shortest interface sequence.
We project these vectors onto the first two text principal components using the same text mean and axes for all interfaces.
The two components explain 16.12\% and 7.64\% of the normalized text-token variance.

\section{Support Size and Computational Cost}
\label{app:geometry}
\label{app:efficiency}

Top-$K$ routing bounds the number of embedding rows used at each position, and the normalized coefficients determine how concentrated each mixture is, with $K-1$ continuous degrees of freedom.
For $T'$ pooled audio positions and routing dimension $d_r$, query--key scoring requires work proportional to $T'Vd_r$, and the subsequent mixture uses $K$ embedding rows per position.

The four Top-$K$ models are evaluated after 18,000 Stage 3 updates with the same tasks and scoring as the speech-input rows of Table~\ref{tab:text-bridge-ablation}.
Table~\ref{tab:app-topk} lists their full results.
All four start Stage 3 from the same Stage 2 checkpoint, which is trained with $K=16$, and use the Stage 3 optimization settings of the \cgate{} model in Table~\ref{tab:text-bridge-ablation}, which is the $K=16$ run.
With $K=1$, the softmax over a single element is constant and gives the router no gradient.
The $K=1$ run therefore uses a straight-through estimator, in which the forward pass uses the highest-scoring embedding and the backward pass uses the gradient of the softmax-weighted average over the vocabulary.

\end{document}